\documentclass[final]{cvpr}
\usepackage{times}
\usepackage{epsfig}
\usepackage{graphicx}
\usepackage{amsmath}
\usepackage{amssymb}

\DeclareMathOperator*{\argmin}{argmin}

\usepackage[pagebackref=true,breaklinks=true,colorlinks,bookmarks=false]{hyperref}
\usepackage{hyperref}
\makeatletter
\newcommand{\printfnsymbol}[1]{%
  \textsuperscript{\@fnsymbol{#1}}%
}
\makeatother
\def\cvprPaperID{****} 
\def\confYear{CVPR 2021}

\begin{document}

\title{3D Meta-Registration: Learning to Learn Registration of 3D Point Clouds}

\author{Lingjing Wang\thanks{Equal contribution} \qquad Yu Hao\printfnsymbol{1} \qquad Xiang Li \qquad Yi Fang\thanks{Corresponding author}\\
NYU Multimedia and Visual Computing Lab\\
    New York University Abu Dhabi, Abu Dhabi, UAE\\
    New York University, New York, USA\\
{\tt\small \{lw1474, yh3252, xl845, yfang\}@nyu.edu}
}

\maketitle

\begin{abstract}
Deep learning-based point cloud registration models are often generalized from extensive training over a large volume of data to learn the ability to predict the desired geometric transformation to register 3D point clouds. In this paper, we propose a meta-learning based 3D registration model, named 3D Meta-Registration, that is capable of rapidly adapting and well generalizing to new 3D registration tasks for unseen 3D point clouds. Our 3D Meta-Registration gains a competitive advantage by training over a variety of 3D registration tasks, which leads to an optimized model for the best performance on the distribution of registration tasks including potentially unseen tasks. Specifically, the proposed 3D Meta-Registration model consists of two modules: 3D registration learner and 3D registration meta-learner. During the training, the 3D registration learner is trained to complete a specific registration task aiming to determine the desired geometric transformation that aligns the source point cloud with the target one. In the meantime, the 3D registration meta-learner is trained to provide the optimal parameters to update the 3D registration learner based on the learned task distribution. After training, the 3D registration meta-learner, which is learned with the optimized coverage of distribution of 3D registration tasks, is able to dynamically update 3D registration learners with desired parameters to rapidly adapt to new registration tasks. We tested our model on synthesized dataset ModelNet and FlyingThings3D, as well as real-world dataset KITTI. Experimental results demonstrate that 3D Meta-Registration achieves superior performance over other previous techniques (e.g. FlowNet3D).
\end{abstract}

\section{Introduction}
The point cloud registration is defined as a process to determine the spatial geometric transformations (i.e. rigid and non-rigid transformation) that can optimally register the source point cloud towards the target one. In comparison to classical registration methods \cite{besl1992method,yang2015go,myronenko2007non}, learning-based registration methods \cite{liu2019FlowNet3D,balakrishnan2018unsupervised} usually leverage a neural network-based structure to directly predict the desired transformation for a given pair of source and target point clouds. 

\begin{figure*}
\centering
\includegraphics[width=12cm]{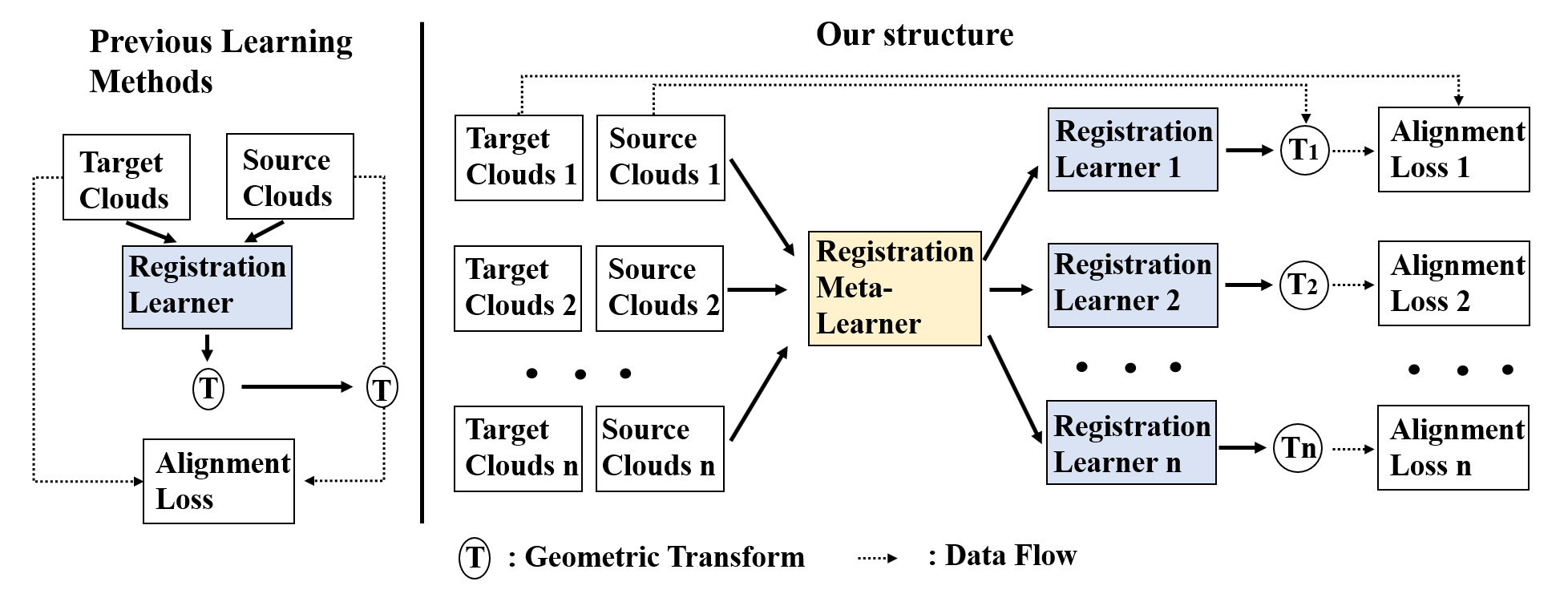}
\caption{Comparison of the pipeline between previous learning methods and our solution for point cloud registration. Previous methods have the 3D registration learner with a fixed structure optimized from minimizing an alignment loss based on data. In comparison, our method treats each input pair as an individual task and adjusts the registration learner accordingly by the registration meta-learner module.}
\label{fig:0}
\end{figure*}

Recently based on the PointNet \cite{qi2017pointnet} structure, Liu et al. proposed FlowNet3D \cite{liu2019FlowNet3D} to learn the points flow field to register two point clouds together. Balakrishnan et al. proposed VoxelMorph \cite{balakrishnan2018unsupervised} for aligning two volumetric 3D shapes. These methods achieved impressive performance for the registration task of 3D shapes/scenes. In comparison to iterative registration methods, learning-based methods have advantages in dealing with a large number of datasets since learning-based methods can transfer the registration pattern from one dataset to another one. However, there are two main challenges for the learning-based methods. Firstly, learning-based networks often require a large volume of data and a long learning period to acquire the ability to predict the desired geometric transformation to register 3D point clouds. Secondly, The generalization capacity can be greatly degraded if the distribution of the dataset in practice differs from the training dataset.  

As shown in Figure \ref{fig:0}, previous learning-based methods treat the ``registration problem'' as a single task for any pair of source and target point clouds. In this way, the pre-trained ``registration learner'' is good at handling the pairs which have similar task distribution to the training pairs. However, in practice, the distribution of registration tasks can be different. Therefore, in comparison to previous learning-based methods, we propose a ``meta-learner'' to treat the registration of each pair of source and target point clouds as an individual registration task. Instead of learning the distribution over data, our meta-learner enables us to learn over the distribution of tasks. After training, the 3D registration ``meta-learner'' is capable of predicting the optimal registration learner to register each pair of source and target point clouds as an independent task. 

\begin{figure*}
\centering
\includegraphics[width=11cm]{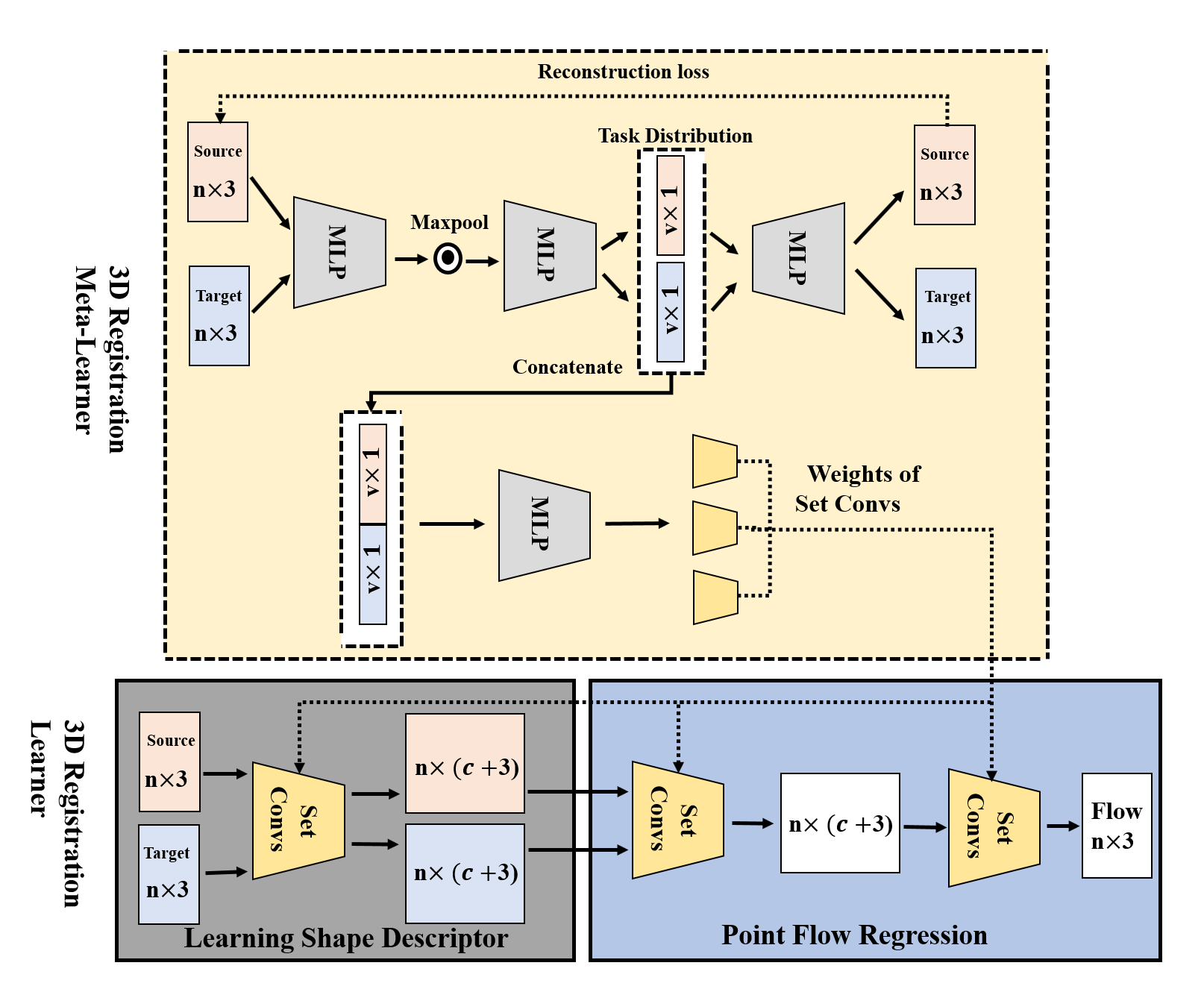}
\caption{Main Pipeline. Our model includes two networks: Registration Learner and Registration Meta-Learner. The registration learner includes a module of learning shape descriptor and points flow regression. The weights in set convolution are predicted from the registration meta-learner.}
\label{fig:main}
\end{figure*}

As shown in Figure \ref{fig:main}, our model includes two modules: 3D registration learner and 3D registration meta-learner. The 3D registration learner includes a module of learning shape descriptor and point flow regression. Other than previous methods, the weights of all the set convolutions in the 3D registration learner are predicted from the 3D registration meta-learner based on the defined task distribution. Therefore, for a given pair of source and target point clouds, the registration learner's structure is updated by the registration meta-learner according to the new registration task. Our contributions are listed as below:

\begin{itemize}
\item In this paper, to the best of our knowledge, it is the first time to introduce a meta-learning strategy for training a model to address challenging issues in 3D registration/mapping/scenes flowing problems in 3D computer vision.

\item In this paper, with the meta-learning training paradigm, our proposed 3D Meta-Registration approach introduces and defines the novel concept of the ``3D registration task'' which differentiates our method from conventional learning-based registration approaches. Our 3D Meta-Registration gains a competitive advantage by training over a variety of the 3D registration task, which leads to an optimized model for the best performance on the distribution of registration tasks, instead of the distribution of 3D data.
 
\item In this paper, we compared our 3D Meta-Registration to other state-of-the-art ones on widely used benchmark datasets (i.e. KITTI, FlyingThings3D) and demonstrated superior registration performance over both seen and unseen data. 
 
\end{itemize}

\section{Related Works}
\subsection{Point Cloud Registration}
In comparison to classical methods \cite{besl1992method,yang2015go,myronenko2007non}, learning-based methods have significant advantages in dealing with a large number of datasets by transferring the ``knowledge'' from registering training pairs to testing pairs. Based on the feature learning structure proposed by PointNet \cite{qi2017pointnet}, Aoki et al. \cite{aoki2019pointnetlk} proposed PointNetLK for rigid point cloud registration by leveraging Lucas $\&$ Kanade algorithm.  Based on DGCNN \cite{wang2019dynamic}, Wang et al. proposed Deep Closest Point \cite{wang2019deep} for learning rigid point cloud registration and PR-Net \cite{wang2019prnet} for learning partial shapes registration. For non-rigid point cloud registration, Liu et al. proposed FlowNet3D \cite{liu2019FlowNet3D} to learn the points flow field for non-rigid point cloud registration. Wang et al. proposed FlowNet3D++ \cite{wang2020FlowNet3D++} on top of FlowNet3D by adding geometric constraints and angular alignment to dramatically improve the alignment performance. However, for most previous methods, the learning structure is a ``fixed'' term after training, which dramatically increases the difficulty to register testing pairs in one forward pass. In comparison, we firstly propose a novel meta-learning-based approach for learning to learn the non-rigid point cloud registration with a flexible and adaptive registration network. The structure of our model can be adjusted accordingly based on the distribution of input point clouds.

\subsection{Meta-learning methods}
Meta-learning \cite{andrychowicz2016learning, schmidhuber1992learning, hospedales2020meta} refers to a subfield of machine learning that learning new concepts and skills much faster and more efficiently given only a small amount of dataset. Parameters prediction \cite{finn2017model, finn2018probabilistic, lee2018gradient} is one of the strategies in meta-learning, which refers to a network trained to predict the parameters of another network so that the first network can encode the related information to the second network which makes the overall network more flexible and adaptive to a particular task. Recently, meta-learning approaches are widely used in computer vision tasks. MANN was proposed by Santoro et al. \cite{santoro2016meta} to use an explicit storage buffer which is easier for the network to rapidly incorporate new information and not to forget in the future. Ravi et al. \cite{ravi2016optimization} use the LSTM to learn an update rule for training a neural network in few-shot learning. Snell et al. \cite{snell2017prototypical} proposed Prototypical networks for few-shot classification task which map the sample data of each class into a metric space by calculating the euclidean distance of prototype representations of each class. In the field of 3D computer vision, Littwin et al. \cite{littwin2019deep} firstly use a deep neural network to map the input point cloud to the parameters of another network for the 3D shape representation task. Yang et al. \cite{yang2020meta3d} proposed Meta3D which an external memory is used to store image features and their corresponding volumes. In this paper, we first propose a meta-learning based method with a 3D registration meta-learner, which can learn to learn the registration pattern more efficiently. 

\section{Methods}\label{headings}
We introduce our approach in the following sections. In section \ref{set1}, we state the problem of learning-based registration. Section \ref{set2} illustrates the 3D registration learner. The 3D registration meta-learner is explained in section \ref{set3}. The definition of the loss function is discussed in section \ref{set4}.

\subsection{Problem Statement}\label{set1}
For a given training dataset $\bold{D}=\{(S_i, G_i)\}) \text{ ,where } S_i, G_i \subset \mathbb{R}^3 \}$, where $S_i$ is the source point cloud and $G_i$ is the target point cloud, we need to define the optimization task. We assume the existence of a parametric function $g_{\theta}(S_i,G_i) = \phi$ using a neural network structure, where $\phi$ is the transformation function (point flow in this paper) which deforms source point cloud towards the target point cloud. We call $g_{\theta}$ 3D registration learner in this paper and $\theta$ is the set of weights in the 3D registration learner. For previous learning-based network structure, the $\theta$ is optimized using stochastic gradient descent based algorithm for a given dataset:

\begin{equation}
\begin{split}
\bold{\theta^{optimal}} =\argmin_{\theta}[\mathbb{E}_{(S_i,G_i)\sim \bold{D}}[\mathcal{L}(S_i,G_i, g_{\theta}(S_i,G_i))]],
\end{split} 
\end{equation}

, where $\mathcal{L}$ represents a similarity measure. 

In comparison, we assume that the 3D registration learner $g$ includes two sets of parameters: $\theta_1$ and $\theta_2$. $\theta_1$ is pre-trained from training dataset, but $\theta_2$ is predicted by another parametric function $f_\sigma$ which is called 3D registration meta-learner in this paper. Similarly, we have the desired transformation function $\phi=g_{(\theta_1,\theta_2)}(S_i, G_i)$ and we have $\theta_2 = f_\sigma(S_i,G_i)$. For a given training data set, we have:

\begin{equation}
\begin{split}
\bold{\theta_1^{optimal},\sigma^{optimal}} =\\
\argmin_{\theta_1, \sigma}[\mathbb{E}_{(S_i,G_i)\sim \bold{D}}[\mathcal{L}(S_i,G_i, g_{(\theta_1, f_\sigma(S_i,G_i))}(S_i,G_i))]].
\end{split}
\end{equation}

\subsection{3D Registration Learner}\label{set2}
The 3D registration learner includes two modules: learning shape descriptor (\ref{s1}) and point flow regression (\ref{s2}). We discuss them in the following subsections. 

\subsubsection{Learning Shape Descriptor}\label{s1}
For a given pair of input point clouds, we firstly learn to extract the shape features that captures representative and deformation-insensitive geometric information. Let $(S_i,G_i)$ denotes the source and target point clouds and $S_i,G_i \subset \mathbb{R}^3$. $\forall x\in S_i,$ we denote the feature of $x$ as $l_x \in \mathbb{R}^c$. Following the recent architecture from PointNet++ \cite{qi2017pointnet++} and FlowNet3D \cite{liu2019FlowNet3D}, the first set convolution $g_1 : \mathbb{R}^3 \to \mathbb{R}^c$ is a non-linear MLP-based function. We note the pre-trained weights in $g_1$ as $\theta_1^1$ and the meta-learned weights in $g_1$ as $\theta_2^1$. The weights for $g_1$ is element-wise summation of $\theta_1^1$ and $\theta_2^1$. We have:

\begin{equation}
\begin{split}
l_{x}= \text{Maxpool} \{g_1(x_j)\}_{||x_j-x||\le r \wedge x_j \in S_i}
\end{split}
\end{equation}
, where r is a predefined distance and Maxpool is an element-wise max pooling function. 

$\forall x\in S_i$, we further concatenate the coordinates $x$ with the learned feature $l_x$ and we denote it as $[x, l_x] \in \mathbb{R}^{(c+3)}$. Similarly, we can learn the feature of each point in the target point cloud $G_i$. The shape descriptor for source point cloud $S_i$ is: $\{ [x,l_x]\}_{x\in S_i}$ and the shape descriptor for target point cloud $G_i$ is: $\{ [x,l_x]\}_{x\in G_i}$.

\subsubsection{Point Flow Regression} \label{s2}
Based on the learned shape descriptors for both source point cloud $\{ [x,l_x]\}_{x\in S_i}$  and target point cloud $\{ [x,l_x]\}_{x\in G_i}$ from previous section, in this section we introduce two more set convolution structures for point flow regression. We define the second set convolution $g_2: \mathbb{R}^{(2c+9)} \to \mathbb{R}^{(c+3)}$ to learn the relation information between descriptors of source and target point clouds. $g_2$ is a non-linear MLP-based function. We note the pre-trained weights in $g_2$ as $\theta_1^2$ and the meta-learned weights in $g_2$ as $\theta_2^2$. The weights for $g_2$ is the element-wise summation of $\theta_1^2$ and $\theta_2^2$. $\forall x\in S_i$, we denote relation tensor $p_x$ as:

\begin{equation}
\begin{split}
p_{x}= \text{Maxpool} \{g_2([x,l_x,y_j,l_{y_j},x-y_j])\}_{||y_j-x||\le r \wedge y_j \in G_i}
\end{split}
\end{equation}
, where [,] denotes concatenation. 

Based on the learned relation feature $\{ [x,p_x]\}_{x\in S_i}$ from source and target descriptors, we define the third set convolution $g_3: \mathbb{R}^{(c+3)} \to \mathbb{R}^3$ to learn the point flow for each point in the source point cloud. $g_3$ is a non-linear MLP-based function. We note the pre-trained weights in $g_3$ as $\theta_1^3$ and the meta-learned weights in $g_3$ as $\theta_2^3$. The weights for $g_3$ is the element-wise summation of $\theta_1^3$ and $\theta_2^3$. $\forall x\in S_i$, we have the flow $v_x$ as:

\begin{equation}
\begin{split}
v_{x}= g_3([x,p_x])
\end{split}
\end{equation}

Therefore, the transformed source shape $S_i' = \{x+v_x\}_{x\in S_i}$. The weights $\theta_1^1, \theta_1^2, \theta_1^3$ are pre-trained from training process. The weights $\theta_2^1, \theta_2^2, \theta_2^3$ are meta-learned from the 3D registration meta-learner in next section. 

\subsection{3D Registration Meta-Learner} \label{set3}
The 3D registration meta-learner includes two parts: task distribution learning module (\ref{set33}) and network of network module (\ref{set34}). We discuss them in the following subsections. 

\subsubsection{Task distribution learning} \label{set33}
We leverage a variational auto-encoder (VAE) network to learn the task distribution from each pair of input shapes. Instead of learning the detailed features of 3D shapes, we learn the most general global shape information by reducing the dimension of the learned latent distribution space via VAE. More specifically, we use a multi-layer MLP-based function $f_1: \mathbb{R}^3 \to \mathbb{R}^v$ with a max-pooling function to learn the mean and variance of the task distribution for source and target point clouds. We denote the weights in $f_1$ as $\sigma_1$. We denote $(\bold{\mu_{S_i}},\bold{\mu_{G_i}})$ as the mean of task distribution space, and $(\bold{\sigma_{S_i}},\bold{\sigma_{G_i}})$ as the standard deviation of task distribution space.  $\forall (S_i,G_i)$, 
\begin{equation}
\begin{split}
\bold{\mu_{S_i}},\bold{\sigma_{S_i}}= \text{Maxpool} \{ f_1(\bold{x_i})\}_{\bold{x_i}\in \bold{S_i}}
\end{split}
\end{equation}
\begin{equation}
\begin{split}
\bold{\mu_{G_i}},\bold{\sigma_{G_i}}= \text{Maxpool} \{ f_1(\bold{x_i})\}_{\bold{x_i}\in \bold{G_i}}
\end{split}
\end{equation}

For $S_i$ and $G_i$, we sample $\bold{L_{S_i}} \sim N(\bold{\mu_{S_i}},\bold{\sigma_{S_i}}), \bold{L_{G_i}} \sim N(\bold{\mu_{G_i}},\bold{\sigma_{G_i}})$, and $\bold{L_{S_i}}, \bold{L_{G_i}} \subset \mathbb{R}^v $. We further use a multi-layer MLP-based decoder $g: \mathbb{R}^v \to \mathbb{R}^{(n\times3)}$ , where n is the number of points in the input shapes, to reconstruct the input shapes. We have:

\begin{equation}
\begin{split}
\bold{S_i}^{rec}= g(\bold{L_{S_i}})
\end{split}
\end{equation}
\begin{equation}
\begin{split}
\bold{G_i}^{rec}= g(\bold{L_{G_i}})
\end{split}
\end{equation}

\subsubsection{Network of network} \label{set34}
To enable the 3D registration meta-learner to learn the pattern of registration, we introduce a second multi-layer MLP based architecture $f_2: \mathbb{R}^v \to \mathbb{R}^w$, where $w$ indicates the dimension of all the weights included in $\theta_2^1, \theta_2^2, \theta_2^3$. We denote the weights in $f_2$ as $\sigma_2$. The meta-learned weights in the 3D registration learner are predicted from the following. $\forall (S_i,G_i)$, we have:
\begin{equation}
\begin{split}
\theta_2^1, \theta_2^2, \theta_2^3=f_2([\bold{L_{S_i}},\bold{L_{G_i}}])
\end{split}
\end{equation}

Therefore, for any given pair of source and target point clouds, our 3D registration learner can be accordingly adjusted by adding the meta-learned weights of $\theta_2^1, \theta_2^2, \theta_2^3$ from the 3D registration meta-learner together with the pre-trained weights of $\theta_1^1, \theta_1^2, \theta_1^3$. The weights $\sigma_1$ and $\sigma_2$ are learned from training dataset with $\theta_1^1, \theta_1^2, \theta_1^3$. 

\subsection{Loss function} \label{set4}
For the target point cloud $G_i$ and transformed source point cloud $S_i'$, we define the loss function in this section. Assuming that we have ground truth flow field $\{v_x^*\}_{x\in S_i}$, we use the simple $L_1$ loss between predicted flow field and ground truth flow field with a cycle consistency term as regularization. We note $v_x'$ as the predicted flow from $S_i'$ to $S$. The loss function is defined as:
\begin{equation}
\begin{split}
\mathcal{L}(S_i',G_i)=\frac{1}{|S_i'|}\sum_{x\in S_i}\{||v_x-v_x^*|| + \lambda ||v_x+v_x'|| \}
\end{split}
\end{equation}
, where $|*|$ denotes the set cardinality and $\lambda$ is a pre-defined hyper-parameter to balance the two terms.  $\lambda$ is set to 0.3 in this paper. 

For the reconstruction loss and unsupervised alignment loss, we leverage the Chamfer distance as the loss function. We formulate the regularized Chamfer loss between our reconstructed point set $S_i^{rec}, G_i^{rec}$ and input points set $S_i, G_i$ as:
\begin{equation} 
\begin{split}L_{\text{Chamfer}}(S_i^{rec},S_i)
 = \sum_{x\in S_i^{rec}}\min_{y \in S_i}||x-y||^2_2\\
 + \sum_{y\in S_i}\min_{x \in S_i^{rec}}||x-y||^2_2 + \lambda KL(N(\mu_{S_i},\sigma_{S_i}), N(0,1))
\end{split}
\end{equation}

\begin{equation} 
\begin{split}L_{\text{Chamfer}}(G_i^{rec},G_i)
 = \sum_{x\in G_i^{rec}}\min_{y \in G_i}||x-y||^2_2\\
 + \sum_{y\in G_i}\min_{x \in G_i^{rec}}||x-y||^2_2 + \lambda KL(N(\mu_{G_i},\sigma_{G_i}), N(0,1))
\end{split}
\end{equation}
, where $\lambda$ is the balance term for the KL divergence regularization.


\section{Experiment}
\label{others}
In this section, we describe the dataset and experimental settings in section \ref{exp1} and section \ref{exp2}. In section \ref{exp3} we perform an oblation study on ShapeNet and Stanford3D dataset to demonstrate our model's performance on unseen categories. In section \ref{exp4}, we compare our model with previous methods using FlyingThings3D and KITTI to demonstrate our model's generalization ability on unseen real Lidar scans. 

\subsection{Dataset preparation} \label{exp1}

\noindent{\textbf{ModelNet40:}}  This dataset contains 12311 pre-processed CAD models from 40 categories. For each 3D point object, we uniformly sample 2048 points from its surface. For each source shape $S_i$ we generate the transformed shapes $G_i$ by applying a rigid transformation defined by the rotation matrix which is characterized by 3 rotation angles along the x-y-z-axis, where each value is uniformly sampled from $[0,45]$ unit degree, and the translation which is uniformly sampled from $[-0.5, 0.5]$. We re-scale both source and target point sets to $[-10,10]$ for a comparison with other datasets.

\noindent{\textbf{ModelNet40 with noise:}} For the testing dataset of ModelNet40, we prepare three types of noise added on both source and target point set as shown in Figure \ref{f1_2}. To prepare the position drift (P.D.) noise, we applied a zero-mean and 0.05 standard deviation Gaussian to each sample from the point set. To prepare the data incompleteness (D.I.) noise, we randomly choose a point and keep its nearest 1536 points from both the source and target point set. To prepare the data outlier (D.O.) noise, we randomly add 148 points as outliers, sampled from a $[-10,10]\times[-10,10]\times[-10,10]$ uniform distribution.

\noindent{\textbf{FlyingThings3D:}} The FlyingThings3D dataset is an open-source collection, which consists of more than 39000 stereo RGB images with disparity and optical flow ground truth. By following the process procedures provided by FlowNet3D, we generate 3D point clouds and registration ground truth using the disparity map and optical map rather than using RGB images. 

\noindent{\textbf{KITTI:}} Another dataset used in this paper is the KITTI scene flow dataset, which consists of 200 training scenes and 200 test scenes. Following previous work FlowNet3D, we use the pre-processed point clouds data which is generated using the original disparity map and ground truth flow.

\subsection{Experimental settings} \label{exp2}
In our model, the batch size is set to 16 and we use Adam optimizer as our optimizer. For the 3D registration learner, the first set convolution includes 6 MLPs with dimensions of (32, 32, 64, 64, 64, 128). The second set convolution includes 6 MLPs with dimensions of (128, 128, 256, 256, 256, 512). The third set convolution includes 6 MLPs with dimensions of (256,256,128,128,256,256) and 2 fully connected layers with dimensions of (128,3). For the 3D registration meta-learner, we use 3 MLPs with dimensions of (32, 32, 64) with a max-pooling layer and one fully connected layer with dimension (64, 128) for learning the mean of VAE and one fully connected layer with dimension (64,128) for learning the standard deviation of VAE. Then we use a fully connected layer with dimensions of (128, 2048$\times$3) for reconstruction. Then we have three fully connected layers with dimensions of 128, 2048, 2048, and the number of all parameters in the set convolution layers. We use the ReLU activation function and implement batch normalization for every MLP layer except the last output layer. We set the learning rate as 0.001 with exponential decay of 0.7 at every 20000 steps.

For evaluation of point cloud registration performance, we use 3D endpoint error (EPE) and point cloud registration estimation accuracy (ACC) with different thresholds as introduced in FlowNet3D. The 3D EPE measures the L2 distance between the point cloud registration estimation and ground truth. A lower EPE value indicates a better estimation of the point flow field. ACC measures the portion of the points with estimated flow field error which is less than one selected threshold among all the points.

\begin{table}[h]
  \caption{Results on the four testing categories for different settings of number of dimension (v) of task distribution. C.D. stands for Chamfer Distance.}
  \label{t1_1}
  \small
  \centering
  \begin{tabular}{lllll}
    \hline               
    Settings     & EPE     & ACC (0.05)     & ACC (0.1) & C.D.\\
    \hline
    v=32 & 0.3003  & 21.43$\%$  & 62.23$\%$   & 0.1291    \\
    \hline
    v=64     & 0.2720  & 25.48$\%$  & 66.41$\%$& 0.1158  \\
    \hline
    v=128     & \bf0.2342  & \bf34.01$\%$  & \bf75.56$\%$& \bf0.0949\\
    \hline
    v=256     & 0.2907  & 21.25$\%$  & 62.79$\%$&  0.1264 \\
    \hline
    v=512    & 0.2994  & 19.28$\%$  & 60.61$\%$& 0.1290  \\
    \hline
  \end{tabular}
\end{table}

\subsection{Experiments on unseen category dataset} \label{exp3}
In this section, we conduct a series of experiments to verify the effect of our model in the registration of unseen data set from different categories.\\

\begin{figure}[h]
\centering
\includegraphics[width=6.5cm]{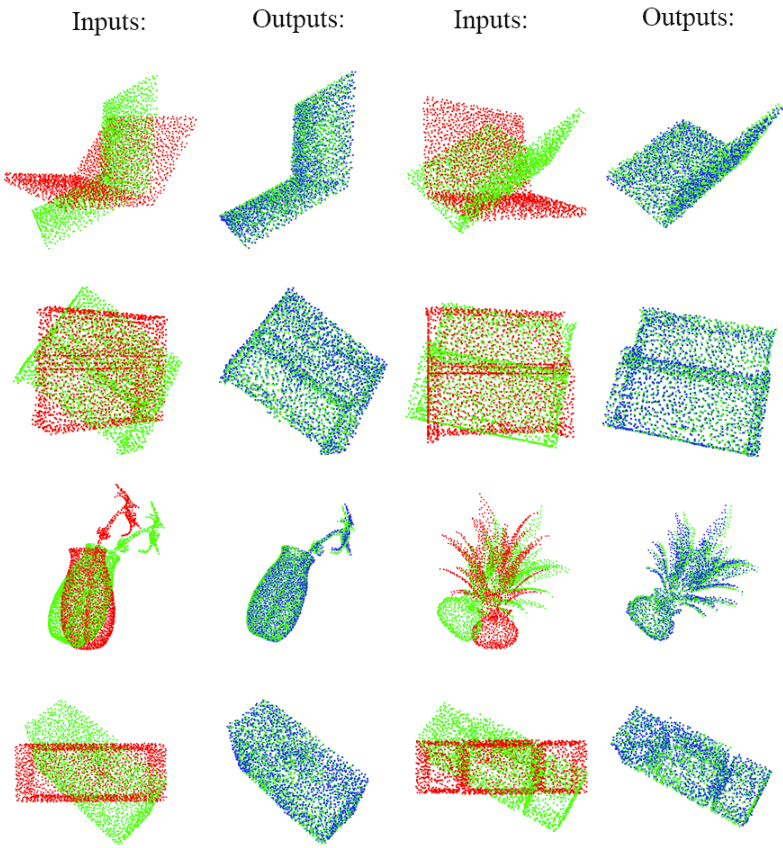}
\caption{Randomly selected qualitative results of point cloud registration on 20 unseen testing categories of ModelNet40 dataset.}
\label{f1_1}
\end{figure}

\begin{figure}[h]
\centering
\includegraphics[width=7.5cm]{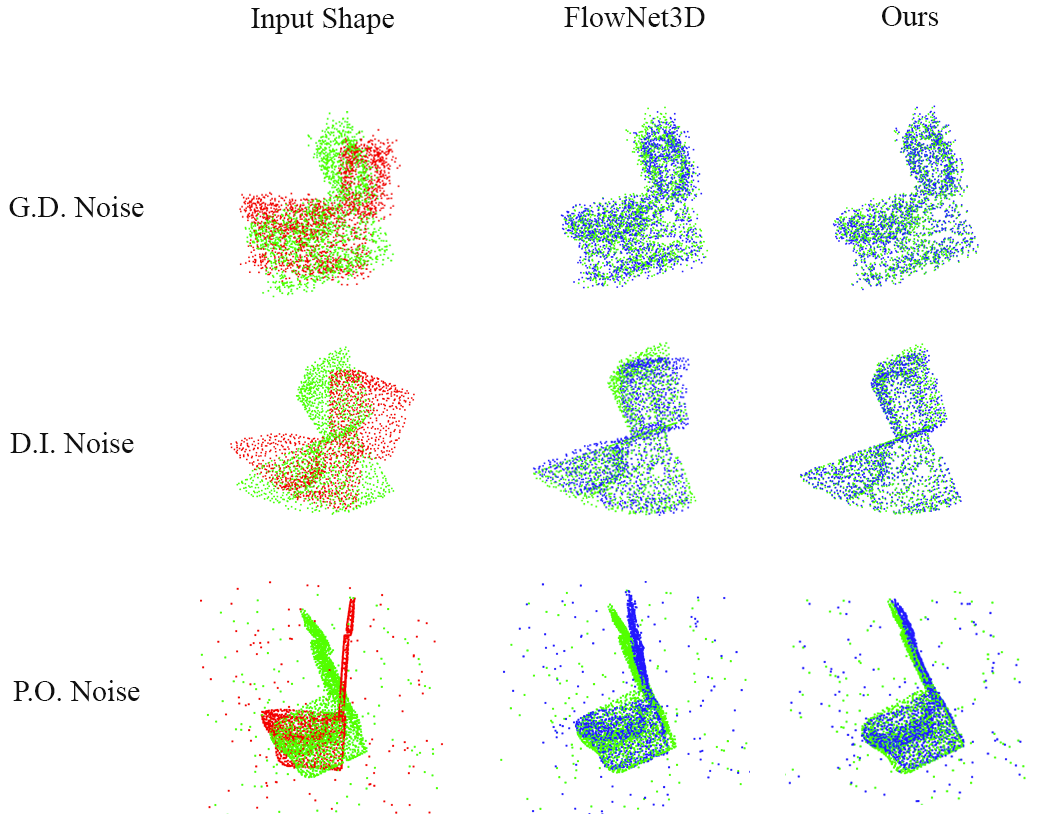}
\caption{Randomly selected qualitative results of point cloud registration on 20 unseen testing categories of ModelNet40 with noise dataset.}
\label{f1_2}
\end{figure}

\noindent{\textbf{Experiment setting:}} We use first 20 categories in ModelNet40 dataset as our training dataset. We use the last 20 categories in ModelNet40 as our testing dataset. For Test 3, we use the last 20 categories in ModelNet40 with noise dataset as our testing data set. For Test 1, we test the effects of dimension $v$ introduced in section \ref{set33} as the dimension of learned task distribution. For Test 2 and 3, we compare our model with meta-learner and without meta-learner. Our model without meta-learner is the final version of FlowNet3D (F3D). We do exact fair comparison for all the following three Tests to show the effect of our proposed meta-learner in dealing with unseen dataset. We list the quantitative results for Test 1 in Table \ref{t1_1}, for Test 2 in Table \ref{t1_2} and for Test 3 in Table \ref{t1_3}. We demonstrate selected qualitative results for Test 2 in Figure \ref{f1_1} and results for Test 3 in Figure \ref{f1_2}. 

\begin{table}[h]
  \caption{Results on the 20 unseen testing categories of ModelNet40 dataset.}
  \label{t1_2}
  \centering
  \small
  \begin{tabular}{llll}
    \hline
    Method     & EPE     & ACC (0.05)     & ACC (0.1) \\
    \hline
    F3D & 0.3648  & 10.01$\%$  & 45.18$\%$    \\
    Ours & \bf0.2342  & \bf34.01$\%$  & \bf75.56$\%$      \\
    \hline
  \end{tabular}
\end{table}

\begin{table}[h]
  \caption{Results on the 20 unseen testing categories of ModelNet40 dataset with noise.}
  \label{t1_3}
  \centering
  \small
  \begin{tabular}{llll}
    \hline
    Method     & EPE     & ACC (0.05)     & ACC (0.1) \\
    \hline
    (G.D.Noise) F3D & 0.4392  & 6.75$\%$  & 34.47$\%$    \\
    (G.D.Noise) Ours & \bf0.3128  & \bf20.92$\%$  & \bf61.30$\%$      \\
    \hline
    (D.I.Noise) F3D & 0.3749  & 11.67$\%$  & 46.07$\%$    \\
    (D.I.Noise) Ours & \bf0.2689  & \bf24.04$\%$  & \bf65.84$\%$      \\
    \hline
    (P.O.Noise) F3D & 0.5738  & 4.81$\%$  & 24.05$\%$    \\
    (P.O.Noise) Ours & \bf0.4156  & \bf10.49$\%$  & \bf40.91$\%$      \\
    \hline
  \end{tabular}
\end{table}

\begin{figure*}
\centering
\small
\includegraphics[width=1.0\linewidth]{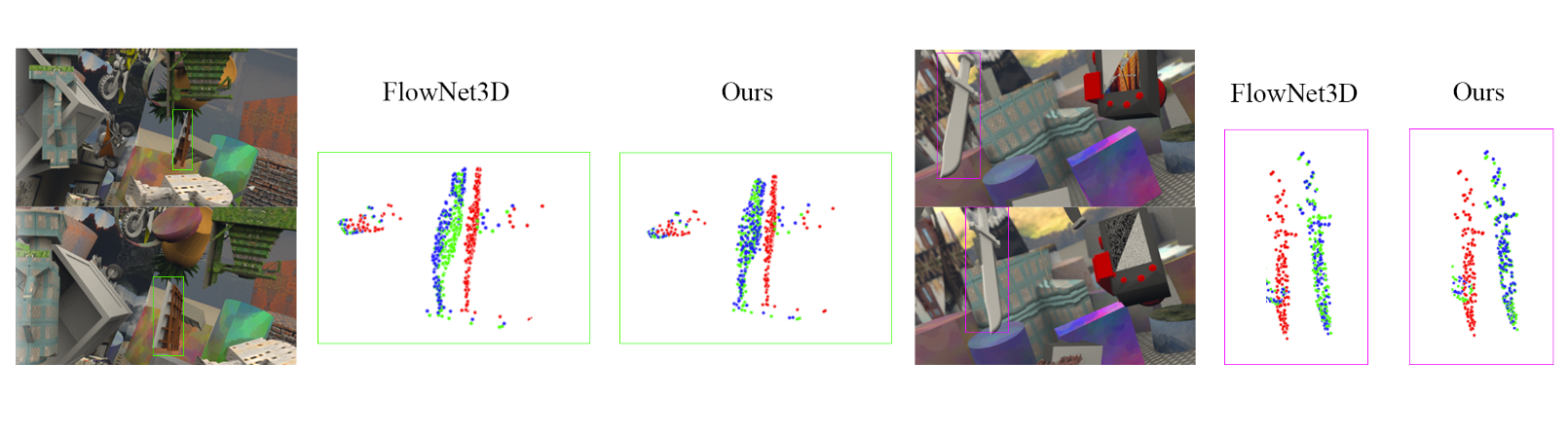}
\caption{Selected point cloud registration qualitative results on the FlyingThings3D dataset. Red points represent the source point cloud. Green points represent the target point cloud. Blue points represent the transformed source point cloud.}
\label{fig:2}
\end{figure*}
\noindent{\textbf{Results of Test 1:}} As shown in Table \ref{t1_1}, we show that our model achieves the best performance on testing unseen categories when the dimension of hidden latent task distribution space is equal to 128. As dimension v is comparatively small, only the most general shape information can be captured and embedded into a latent vector as the task distribution. Therefore, similar shapes from different categories may have close task distribution, and the predicted weights of registration learner can be similar to each other as well to generate a good performance on the unseen categories dataset. As v increases from 128 to 512, more detailed and subtle shape information is embedded into the latent vector. The general task distribution tends to converge to detailed data distribution. The over-fitted model can not be generalized on unseen categories. In this case, the model's performance degrades accordingly on the unseen categories dataset. On the other hand, when the dimension of hidden latent task distribution space v is too small (for example, v is equal to 32), different tasks cannot be distinguished based on the latent vectors and the model's performance tends to degrade towards the baseline model (F3D) without meta-learner.

\noindent{\textbf{Results of Test 2:}}
As shown in Table \ref{t1_2}, with the meta-learner our model achieves much lower EPE (0.2342) compared to the model without meta-learner (F3D) on the 20 unseen testing categories. The registration estimation accuracy (ACC) dramatically improves from10.01$\%$ to 34.01$\%$ for the threshold 0.05 and it improves from 48.18$\%$ to 75.56$\%$ for the threshold 0.1. This result clearly indicates the superior performance of leveraging meta-learner to adjust the learner based on the task distributions of unseen categories. In addition, from the randomly selected qualitative results from Figure \ref{f1_1}, we notice that the alignment result of source and target point clouds by our model is precise. The green shape is the target, the red shape is the source shape and the blue one is the transformed source shape. 

\noindent{\textbf{Results of Test 3:}} As shown in Table \ref{t1_3}, for all these three types of noise (G.D, D.I, and P.O noise), with the meta-learner our model achieves much lower EPE and higher accuracy compared to the model without meta-learner (F3D) on the unseen 20 testing categories. This result indicates that our model with meta-learner can be less affected by all these different noise patterns in comparison with the model without meta-learner (F3D). For more reference, we list one qualitative results for each noise pattern in Figure \ref{f1_2} to further show the difference in alignment results by our model with meta-learner with baseline model FlowNet3D (F3D) without meta-learner. We can clearly see that our model achieves better alignment performance. 

\begin{table}[h]
  \caption{Results on the FlyingThings3D test dataset.}
  \label{tt3}
  \centering
  \small
  \begin{tabular}{llll}
    \hline
    Method     & EPE     & ACC (0.05)     & ACC (0.1) \\
    \hline
    ICP & 0.5019  & 7.62$\%$  & 7.62$\%$    \\
    FlowNet3D (EM)  & 0.5807  & 2.64$\%$  & 12.21$\%$      \\
    FlowNet3D (LM)  & 0.7876  & 0.27$\%$  & 1.83$\%$      \\ 
    FlowNet3D & 0.1694  & 25.37$\%$  & 57.85$\%$      \\
    \hline
    Ours     & \bf0.1453  & \bf29.27$\%$  & \bf62.21$\%$   \\
    \hline
  \end{tabular}
\end{table}

\begin{figure*}
\centering
\includegraphics[width=10.5cm]{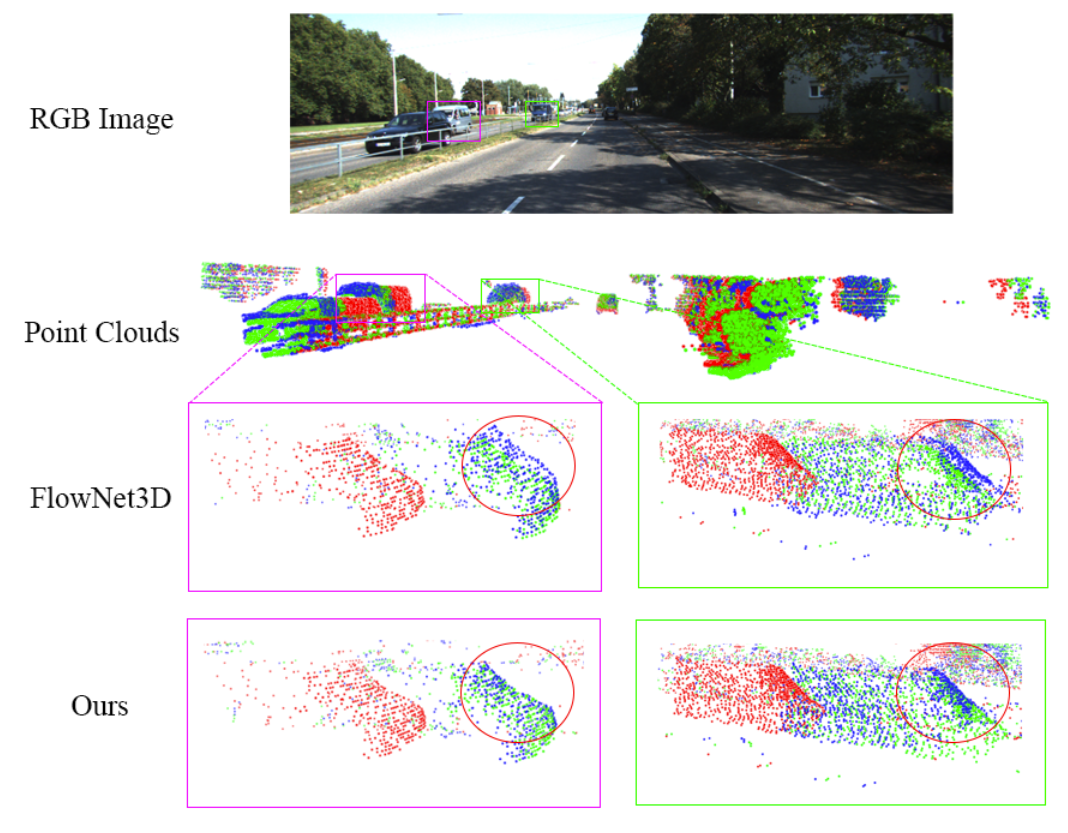}
\caption{Qualitative results of point cloud registration on the KITTI dataset. Red points represent the source point cloud. Green points represent the target point cloud. Blue points represent the transformed source point cloud.}
\label{fig:3}
\end{figure*}

\subsection{Experiments on unseen real-world dataset} \label{exp4}
In this experiment, to better demonstrate the generalization capacity of our model, we train our model using synthesized dataset FlyingThings3D and evaluate our model's performance on real Lidar scan KITTI dataset for 3D point cloud registration.

\paragraph{Experiment setting:} We conduct two Tests in this section. For both Test 1 and Test 2, we from nearly 39000 models in the FlyingThings3D dataset, we randomly select 20000 models for training. For Test 1, we use the rest of the 2000 models as the testing dataset. We notice that Test 1 is conducted on a ``seen'' testing dataset to compare with other methods. We sample 2048 points from each point cloud. In Test 1, we compare our model with ICP (iterative closest point) and three settings of FlowNet3D to show our model's performance on a testing dataset from the same domain. FlowNet3D refers to the optimal model in \cite{liu2019FlowNet3D}. FlowNet3D (EM) refers to an early mixture model that combines two point clouds with a one-hot vector to distinguish them. FlowNet3D (LM) denotes a late mixture model that firstly extracts global features from the source and target point cloud respectively and then uses concatenation operation to mix them. Note that we exactly follow the same experiment setting in FlowNet3D for a fair comparison. We train our model on the training dataset and evaluate the performance on the testing dataset. We list the quantitative results for Test 1 in Table \ref{tt3} and we demonstrate selected qualitative results for Test 1 in Figure \ref{fig:2}. 

For Test 2, we only evaluate our model on the KITTI scene flow dataset. The KITTI scene flow dataset includes 200 stereo images. Following the evaluation standards claimed in FlowNet3D, we use the first 150 images containing only geometric information from the KITTI scene flow dataset as the testing dataset to evaluate our model. Note that our model is trained on the training dataset of FlyingThings3D without further fine-tuning steps. We use all 16384 points from each point cloud and set the batch size to 1 during the evaluation process. For Test 2, we compare our model's performance with the model FlowNet3D. We list the quantitative results for Test 2 in Table \ref{tt4}. We demonstrate selected qualitative results for Test 2 in Figure \ref{f1_2}. 

\noindent{\textbf{Results of Test 1:}} As shown in Table \ref{tt3}, our method achieves lower EPE (0.1453) compared to FlowNet3D (0.1694) and ICP (0.5019). As to the registration estimation accuracy (ACC), our method achieves significantly better result with 29.27$\%$ for the threshold 0.05 and 62.21$\%$ for the threshold 0.1, which is better than 26.67$\%$ for the threshold 0.05 and 59.65$\%$ for the threshold 0.1 achieved by FlowNet3D. In addition, from the qualitative results shown in Figure \ref{fig:2}, we notice that the alignment of source and target point clouds is more accurate for our method. For example, we can clearly see the gap between green (target) and blue (transformed source) points for the left case in the result of FlowNet3D.  

\noindent{\textbf{Results of Test 2:}} The results shown in Table \ref{tt4} demonstrates that our model trained on FlyingThings3D achieves lower EPE and better ACC on KITTI in comparison to the results achieved by FlowNet3D. For the qualitative results shown in Figure \ref{fig:3}, we can clearly see that our registration result for the two cars on the left side is better since all the blue points are almost overlapped with the green points. In comparison, the result of FlowNet3D shows a gap between blue and green points. 

\begin{table}
  \caption{Results on the KITTI dataset.}
  \label{tt4}
  \centering
  \begin{tabular}{llll}
    \hline               
    Method     & EPE     & ACC (0.05)     & ACC (0.1) \\
    \hline
    FlowNet3D & 0.2113  & 8.79$\%$  & 32.85$\%$      \\
    \hline
    Ours     & \bf0.1755  & \bf11.27$\%$  & \bf43.25$\%$   \\
    \hline
  \end{tabular}
\end{table}

\section{Conclusion}
This paper introduces a novel meta-learning-based approach to our research community for point cloud registrations. In contrast to recent proposed learning-based methods, our method leverages a 3D registration meta-learner to learn the distribution of registration tasks instead of the distribution of data. For each pair of source and target point clouds, the 3D registration meta-learner can accordingly predict an optimal structure for the 3D registration learner to address the desired transformation for registration. To the best of our knowledge, our method firstly leveraged a meta-learning based structure for this task and we achieved superior point cloud registration results on the unseen dataset. 


{\small
\bibliographystyle{ieee_fullname}
\bibliography{egbib}
}

\end{document}